\documentclass[letterpaper]{article} % DO NOT CHANGE THIS
\usepackage{aaai25}  % DO NOT CHANGE THIS
\usepackage{times}  % DO NOT CHANGE THIS
\usepackage{helvet}  % DO NOT CHANGE THIS
\usepackage{courier}  % DO NOT CHANGE THIS
\usepackage[hyphens]{url}  % DO NOT CHANGE THIS
\usepackage{graphicx} % DO NOT CHANGE THIS
\usepackage{natbib}  % DO NOT CHANGE THIS AND DO NOT ADD ANY OPTIONS TO IT
\usepackage{caption} % DO NOT CHANGE THIS AND DO NOT ADD ANY OPTIONS TO IT
\usepackage{algorithm}
\usepackage{algorithmic}
\usepackage{amsmath,epstopdf,amssymb,color,url,multirow,multicol,verbatim,threeparttable,diagbox,makecell,booktabs}
\newcommand*\samethanks[1][\value{footnote}]{\footnotemark[#1]}
\usepackage{newfloat}
\usepackage{listings}
\DeclareCaptionStyle{ruled}{labelfont=normalfont,labelsep=colon,strut=off} % DO NOT CHANGE THIS
\floatstyle{ruled}
\newfloat{listing}{tb}{lst}{}
\floatname{listing}{Listing}
\title{Unsupervised Region-Based Image Editing of Denoising Diffusion Models}
\author{Zixiang Li\textsuperscript{\rm 1,2}, Yue Song\textsuperscript{\rm 4}, Renshuai Tao\textsuperscript{\rm 1,2}, Xiaohong Jia\textsuperscript{\rm 5}, Yao Zhao\textsuperscript{\rm 1,2,3}\thanks{Corresponding author}, Wei Wang\textsuperscript{\rm 1,2}\samethanks}
\affiliations{
    \textsuperscript{\rm 1}Institute of Information Science, Beijing Jiaotong University\\
    \textsuperscript{\rm 2}Visual Intellgence +X International Cooperation Joint Laboratory of MOE\\
    \textsuperscript{\rm 3}Pengcheng Laboratory, Shenzhen, China\\
    \textsuperscript{\rm 4}University of Trento, Italy\\
    \textsuperscript{\rm 5}Lanzhou Jiaotong University\\
\{zixiangli, yzhao\}@bjtu.edu.cn}

\usepackage{bibentry}
\begin{document}

\maketitle

\begin{abstract}

Although diffusion models have achieved remarkable success in the field of image generation, their latent space remains under-explored. Current methods for identifying semantics within latent space often rely on external supervision, such as textual information and segmentation masks. In this paper, we propose a method to identify semantic attributes in the latent space of pre-trained diffusion models without any further training. By projecting the Jacobian of the targeted semantic region into a low-dimensional subspace which is orthogonal to the non-masked regions, our approach facilitates precise semantic discovery and control over local masked areas, eliminating the need for annotations. We conducted extensive experiments across multiple datasets and various architectures of diffusion models, achieving state-of-the-art performance. In particular, for some specific face attributes, the performance of our proposed method even surpasses that of supervised approaches, demonstrating its superior ability in editing local image properties.
\end{abstract}
\section{Introduction}

Diffusion models have achieved remarkable success in the field of image generation, with notable examples including the Denoising Diffusion Probability Model (DDPM)~\cite{ho2020denoising}, the Denoising Diffusion Implicit Model (DDIM)~\cite{song2020denoising}, and score-based generative models~\cite{song2020score}. Diffusion models have set new benchmarks in various applications such as image generation, video generation, image editing, and inverse problems. When deploying diffusion models for downstream tasks like image editing, additional supervision such as textual descriptions or sketches is typically required, and this often involves the integration of supplementary modules or fine-tuning pre-trained models (e.g., Lora~\cite{hu2021lora}, DiffusionClip~\cite{kim2022diffusionclip}). These techniques are meticulously engineered to minimize undesired alterations to the noise space, thereby avoiding irregular changes or artifacts. The necessity for such additional supervision or training underscores the current limitations in the inherent information available within diffusion models. Nevertheless, as pre-trained diffusion models have demonstrated unexpected versatility across various domains, it becomes imperative to explore and leverage their intrinsic capabilities to fully realize their potential. Therefore, it is more favored to leverage the pre-trained models for unsupervised image editing.

Diffusion models have achieved impressive results in guided generation tasks. The most outstanding one is the use of Contrastive Language-Image Pretraining (CLIP)~\cite{radford2021learning} to convert the user input text prompt into the condition of the diffusion model input, thereby generating high-quality images. GLIDE~\cite{nichol2021glide}, DALL·E 2~\cite{ramesh2022hierarchical}, and Stable Diffusion~\cite{rombach2022high} are some of the best models currently. These models use CLIP to align the text space with the latent space of the diffusion model, so as to achieve text-guided image generation. However, there is still no comprehensive exploration of the properties of the latent space of the diffusion model itself. Current research on the latent space of diffusion models remains a bit under-explored. 

\begin{figure}
    \centering
    \includegraphics[width=1\linewidth]{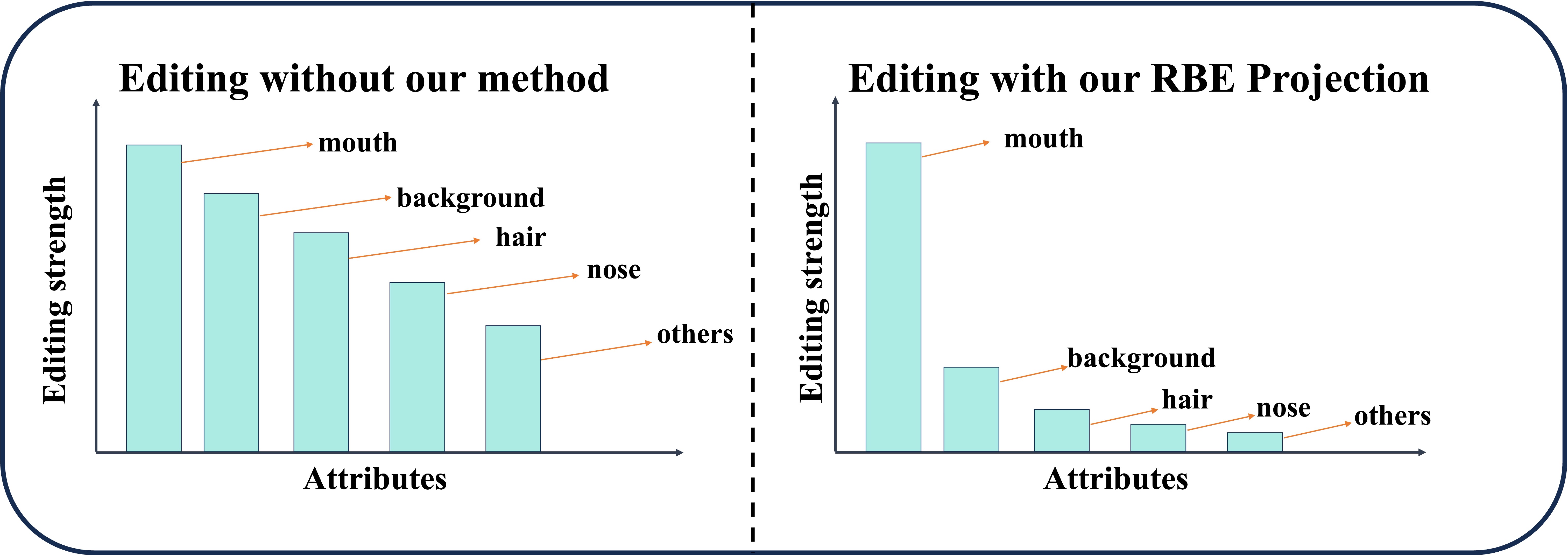}
    \caption{\textbf{Motivation of our proposed method.} In this figure, mouth editing is used as an example. Direct editing often results in significant changes to areas beyond the mouth. By using Jacobian matrix projection, we can suppress these unwanted changes, allowing for more precise editing.}
    \label{fig1:our method}
\end{figure}

In the denoising process of the diffusion model, the image space is generally represented as $\mathbf{x}_{\{1:T\}}$, where $\{1:T\}$ represents the time step of denoising. In the denoising process of the diffusion model, the denoised image space is generally represented as $\mathbf{x}$. All vector dimensions in $\mathbf{x}_{\{1:T\}}$ are the same as the initial input $\mathbf{x}_{0}$ and the output $\epsilon_{\{1:T\}}$ after the denoising network. In this article, we abbreviate the pixel space of the diffusion model as x-space. The x-space within diffusion models is generally considered to be devoid of semantics. To address this, Diffusion autoencoders~\cite{preechakul2022diffusion} introduces a semantic encoder that maps x-space into semantically rich vectors. Subsequently, Kwon~\cite{kwon2022diffusion} proposed that in diffusion models employing the U-Net architecture, the intermediate skip connections, termed h-space, can be utilized to control image semantics. Subsequent investigations~\cite{park2023understanding} have aimed to establish a linkage between h-space and x-space. Despite these efforts, the supervisory information can only be applied within the x-space, and these approaches strive to connect the x-space with the h-space. However, these supervised methodologies fail to fully exploit the potential abilities of identifying inherent semantics in h-space. Subsequent research by Boundary Diffusion~\cite{zhu2024boundary} examined the concurrent alterations in x-space and h-space to facilitate semantic editing. However, their approach relied on simple classification methods to identify vectors controlling semantics, resulting in uncontrollable semantic changes. Semdiff~\cite{haas2024discovering} introduced significant advancements by incorporating masks to locate the local Jacobian matrix. This innovation considerably enhanced the efficiency of identifying local semantics, and as an unsupervised method, it greatly enriched the discovered semantics. Nonetheless, similar to preceding methods, Semdiff focuses solely on local semantic changes and fails to maintain structural details outside the targeted edited area.

In our work, we introduce a novel algorithm named Region-Based Editing (RBE) to identify semantics based on regions while preserving the rest details of the image. Figure~\ref{fig1:our method} shows the core idea of our method. Our method is a fully unsupervised algorithm that neither requires fine-tuning of pre-trained models nor the addition of auxiliary modules. The primary objective is to modify only the pixels within the region of interest, leaving the exterior pixels unchanged. This approach ensures post-editing, where only the targeted region transforms while the surrounding areas remain unaffected. Our pixel-based optimization method concerns decomposing the Jacobian matrix of the noise prediction model, and the goal is to maximize the pixel changes within the region of interest while minimizing changes outside this region. We only need to add a rough bounding box to the area to be edited, apply this box to the noise prediction network, and calculate the Jacobian matrix corresponding to the inside and outside of the box for the modified noise prediction network. Finally, we use the vectors of the area to be edited as a set of bases and use projection to suppress the influence of the vectors represented by the Jacobian matrix outside the area. Our method incorporates information outside the region of interest during local semantic editing within diffusion models. This process requires only a coarse mask, eliminating the need for a fine segmentation network. Moreover, due to the universality and computational efficiency of our framework, it can be applied to all pre-trained diffusion models utilizing the U-Net architecture. Additionally, we investigate the relationship between the h-space of diffusion models and semantic discovery and editing, deepening our understanding of the model's latent space. Extensive experiments and comprehensive evaluations validate the effectiveness of our proposed method.
% 1.semantic in latent space
\section{Background}

\noindent\textbf{Image Editing in Diffusion Model.} In the field of image editing, a large number of works based on diffusion models have emerged in recent years. These works highlight the potential and versatility of diffusion models in improving image editing performance. The image editing of diffusion models has multiple categories, such as text guidance, reference image guidance, semantic segmentation map guidance, and mask guidance, etc. These methods include both earlier traditional and current multimodal conditional methods~\cite{huang2024diffusion}.

Text-based methods are the most commonly used. GLIDE~\cite{nichol2021glide} is the first work to use text to directly control image generation. Unlike image generation tasks, image editing focuses on changing the appearance, content or structure of existing images. DiffusionClip~\cite{kim2022diffusionclip} is an early method that uses CLIP for diffusion models. The success of CLIP has inspired many new explorations~\cite{liu2024forgery,tan2024c2p,huang2023clip2point,jiao2023learning}. For image editing based on diffusion models guided by other conditions, controlnet~\cite{zhang2023adding} is a guidance that adds spatial condition control. \citet{zhang2023controlvideo} further explores control methods in video generation. These works have also had a positive impact on tasks such as continuous learning~\cite{zhu2023ctp,zhang2023slca} and segmentation~\cite{zhang2023coinseg,chen2024saving}.
% Mask-based image guidance is a prevalent method that primarily relies on segmentation networks or manual definitions to apply masks. DiffEdit~\cite{couairon2022diffedit} simplifies the process of semantic editing by automatically generating masks that isolate specific areas for modification. 

Another method is to operate on the latent space of the diffusion models. Their advantage is that the ability to edit images can be achieved without a large amount of paired training data,  which depends on the pre-trained diffusion model. DragDiffusion~\cite{mou2023dragondiffusion} is one of the representatives of the methods, which can directly guide editing through image features. It observes that there is rich semantic information in the features of U-Net, and this information can be used to construct energy functions to guide editing.

Although many methods of image editing have good results in changing content, color, texture, etc., when dealing with complex structures, such as fingers, eyes, corners of the mouth and other details, artifacts are often generated. It is still a challenge to improve the artifact generation during image editing of diffusion models.

\noindent\textbf{Latent Space Disentanglement.} Extensive research has been conducted on the latent space disentanglement of generative models. $\beta$-VAE~\cite{higgins2017beta} introduces a hyperparameter $\beta$ to the KL divergence term, enhancing the gap between prior and posterior distributions to achieve effective disentanglement. Additional unsupervised methods based on statistical features, such as $\beta$-TC-VAE~\cite{kim2018disentangling}, DIP-VAE~\cite{kumar2017variational}, and Guided-VAE~\cite{ding2020guided}, have also demonstrated significant success. Generative Adversarial Networks (GANs)~\cite{goodfellow2014generative} have a well-defined latent space to disentanglement.

Some recent efforts have proposed different approaches to improve the disentanglement of GANs from various perspective~\cite{karras2019style,karras2020analyzing,ling2021editgan,wang2022high}. Furthermore, ReSeFa~\cite{huang2023region} addresses region-based semantic discovery as a dual optimization problem, achieving semantic disentanglement through a properly defined generalized Rayleigh quotient. \citet{song2024flow} use Flow Factorized Representation Learning to achieve more effective structured representation.

Unlike VAES and GANs, the x-space in the diffusion model is considered to be an implicit representation lacking semantics. Diffusion autuencoders~\cite{preechakul2022diffusion} explores the possibility of representation learning of x-space through a semantic encoder. They achieve meaningful encoding and decodability of the input image.
\citet{kwon2022diffusion} further investigate the relationship between the x-space and h-space. Their approaches leverage CLIP to facilitate semantic disentanglement. \citet{zhu2024boundary} and \citet{park2023understanding} explore the relationship between x-space and h-space, achieving a certain degree of disentanglement. \citet{haas2024discovering} identify semantic directions within the h-space. \citet{dalva2024noiseclr} use contrastive learning to achieve unsupervised semantic direction discovery. \citet{wu2024factorized} combine content with mask to discover visual concepts. \citet{hu2024latent} use transformer-based flow matching to achieve latent space editing.

To sum up, modifying local attributes often results in unintended global changes, as current methods focus narrowly on regional alterations, overlooking broader harmony. Moreover, they typically learn only a limited set of concepts. Achieving diverse regional attribute discovery and precise local editing remains a significant challenge.

\section{Methodology}
In this section, we explain the reasons why previous classifier-guided methods alter local attributes while simultaneously affecting global attributes. Subsequently, we introduce our region-based semantic discovery method and a technique for controlling the generation of time-step edits. 

\subsection{Preliminary: Denoising Diffusion Models}

Denoising Diffusion Probability Models (DDPM) is a generative model inspired by the diffusion phenomenon in thermodynamic systems. The original diffusion models do not have a latent space but recently Kwon\cite{kwon2022diffusion} identified a potential semantic latent space at the bottleneck level of U-Net, referred to as h-space. They re-formulated the inverse process of the diffusion model as:
\begin{equation}
    x_{t-1}=\sqrt{\alpha_{t-1}} \mathbf{P}_t(\boldsymbol{\epsilon}_t^\theta(\boldsymbol{x}_t))+\mathbf{D}_t(\boldsymbol{\epsilon}_t^\theta(\boldsymbol{x}_t))+\sigma_t\boldsymbol{z}_t.
\end{equation}
where $\mathbf{z}_t\sim\mathcal{N}(\mathbf{0},\mathbf{I})$ , $\boldsymbol\epsilon_t^\theta$ is a neural network to predict noise from $\mathbf{x}_t$. At the same time, $\mathbf{P}_t(\boldsymbol{\epsilon}_t^\theta(\mathbf{x}_t))$ and $\mathbf{D}_t(\boldsymbol{\epsilon}_t^\theta(\mathbf{x}_t))$ are expressed as:
\begin{equation}
    \mathbf{P}_t(\boldsymbol{\epsilon}_t^\theta(\mathbf{x}_t))=\frac{\mathbf{x}_t-\sqrt{1-\alpha_t}\boldsymbol{\epsilon}_t^\theta(\mathbf{x}_t)}{\sqrt{\alpha_t}}.
\end{equation}
and 
\begin{equation}
    \mathbf{D}_t(\boldsymbol{\epsilon}_t^\theta(\mathbf{x}_t))=\sqrt{1-\alpha_{t-1}-\sigma_t^2}\boldsymbol{\epsilon}_t^\theta(\mathbf{x}_t).
\end{equation}
$\sigma_{t}=\eta\sqrt{(1-\alpha_{t-1})/(1-\alpha_{t})}\sqrt{1-\alpha_{t}/\alpha_{t-1}}$. The range of $\eta$ is from 0 to 1, where 0 corresponds to DDIM and 1 corresponds to DDPM. Its value represents the degree of randomness. During the image editing process of Asyrp, the semantic vector $\Delta\mathbf{h}_t$ is only injected into $\mathbf{P}_{t}$ without changing $\mathbf{D}_{t}$. They proved that the shift caused by $\Delta\mathbf{h}_t$ is offset by the shift $\mathbf{D}_{t}$ caused by $\Delta\mathbf{h}_t$. Therefore, applying $\Delta\mathbf{h}_t$ on $\mathbf{P}_{t}$ and $\mathbf{D}_{t}$ leads to the same output as the original output.

\begin{figure*}[tbp]
    \centering
    \includegraphics[width=0.8\linewidth,height=6.5cm]{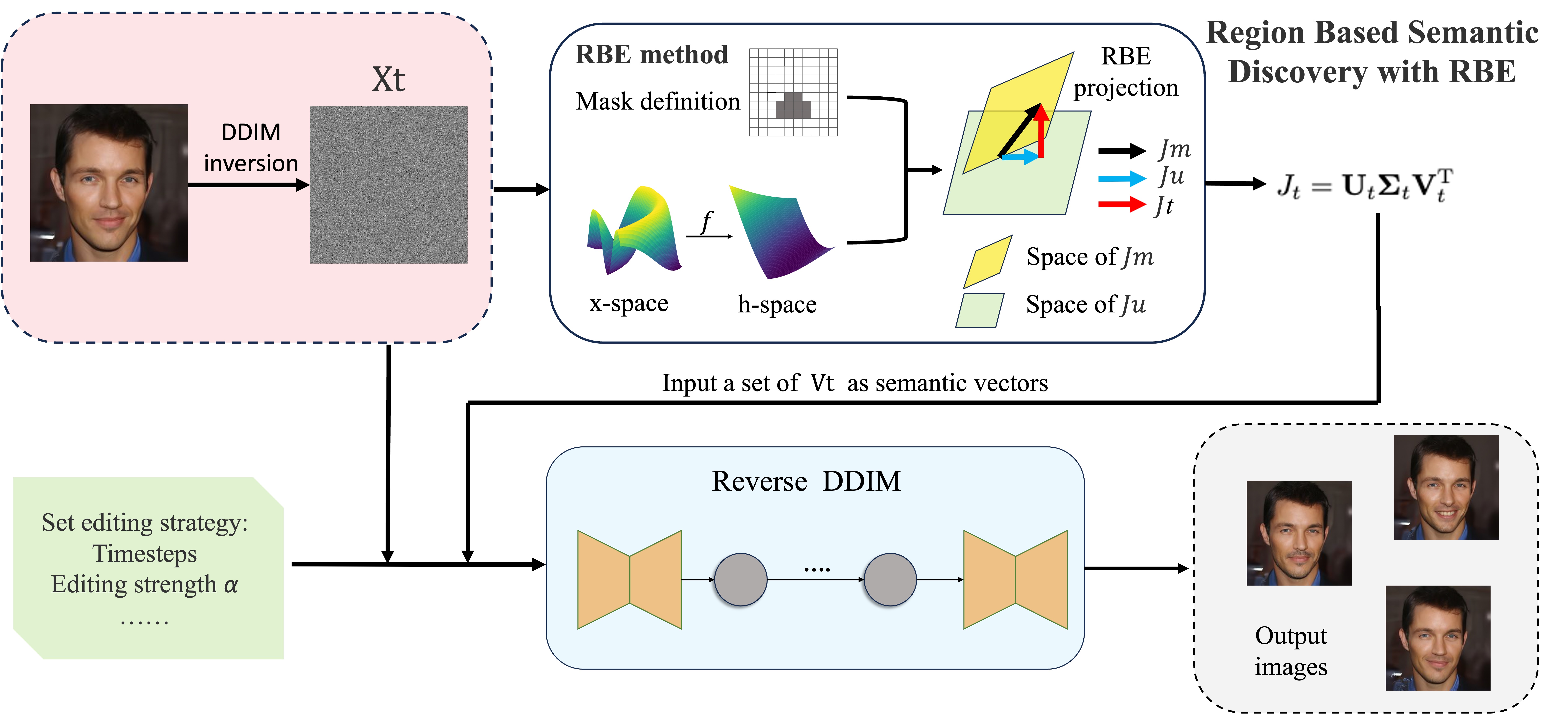}
    \caption{\textbf{Overview of our semantic discovery method and editing method.} Firstly, we define mask M and function $f$, which can be found in our method section. We also use DDIM inversion to precompute $x_t$ and $h_t$ for later use. We use the power iteration method to calculate the Jacobian matrix $J_t$, and use $J_t$ to calculate the required $V_t$. Finally, we set the modified timesteps, edit intensity and other parameters, and use DDIM to generate images. All algorithms and specific experimental settings can be found in the appendix.}
    \label{fig:pipeline}
\end{figure*}
\subsection{Region-Based Semantic Discovery}

We argue that h-space has local Euclidean properties and consistency in different timesteps. This kind of space allows us to propose approaches analogous to the latent disentanglement methods in VAEs and GANs. Therefore, we associate semantic discovery with the matrix of the generative network, thereby achieving semantic discovery that does not require supervision.
% \yue{Unsupervised editing through Jacobian}}

\noindent\textbf{Unsupervised Editing through Jacobian.} In generative models with explicit latent spaces, a semantic direction can usually maximize the output variations, which means a relation between image space and latent space. In diffusion models, if we assume that h-space is a latent space, then we also hope to find the connection between image space and h-space. Given a time step $t$ and the corresponding $\mathbf{x}$, our goal is to find a function that establishes a connection between the denoising network $\mathbf{P}_{t}(\boldsymbol{\epsilon}_{t}^{\theta}(\mathbf{x}_{t}))$ and h-space. To achieve this goal, we define the following function:

\begin{equation}
    f(h_{t}):=\epsilon_{t}^{\theta}(\mathbf{x}_{t},h_t).
\end{equation}
where $\mathbf{x}_{t}$ is determined image in specific timestep. The process of semantic editing can be described as:
\begin{equation}
    x^\mathrm{edit}=f(h_t+\alpha \Delta h_t).
\end{equation}

Inspired by previous work, we relate the direction of maximum semantic change to the Jacobian matrix of the function. The Jacobian matrix is a bridge that plays an important role in attribute editing. The Jacobian matrix is defined as $J_t=\frac{\partial f}{\partial h_t}$.
Descending along the direction of the Jacobian matrix will result in the most significant change in the function, potentially identifying a semantic direction. To analyze this Jacobian matrix, we can perform Singular Value Decomposition (SVD):
% \begin{equation}   
% {J}_t=\mathbf{U}_t\boldsymbol{\Sigma}_t\mathbf{V}_t^\mathrm{T}.
% \end{equation}
\begin{equation}   
{J}_t=\mathit{U}_t\mathit{\Sigma}_t\mathit{V}_t^\mathrm{T}.
\end{equation}
The corresponding right singular vectors $V_{t}$ constitute the orthogonal vector set of the image perturbations in h-space. Finally, our editing method becomes:
\begin{equation}
    f(h_t+\alpha \Delta h_t)\approx f(h_t)+ \alpha V_{t}.
\end{equation}
However, directly computing the Jacobian matrix $J_{t}$ remains computationally expensive. Fortunately, solving the vector product ${J}_t^\mathrm{T}{J}_t$ using power iteration is significantly more feasible. The eigenvectors of ${J}_t^\mathrm{T}{J}_t$ can be accurately approximated to the right singular vectors of $J_{t}$.
Next, we demonstrate why $J_{t}$ can represent the perturbation of h-space and is equivalent to the direction of the maximum change of the image.The known image prediction network pt is expressed as follows:
\begin{equation}
    \mathbf{P}_t(\boldsymbol{\epsilon}_t^\theta(\mathbf{x}_t))=\frac{\mathbf{x}_t-\sqrt{1-\alpha_t}\boldsymbol{\epsilon}_t^\theta(\mathbf{x}_t)}{\sqrt{\alpha_t}}.
\end{equation}
Under the premise that xt is fixed, if $\mathbf{P}_t$ is partial derivative with respect to ht, the result is:
\begin{equation}
\frac\partial{\partial\mathbf{h}_t}\mathbf{P}_t(\mathbf{x}_t,\mathbf{h}_t)=-\frac{\sqrt{1-\alpha_t}}{\sqrt{\alpha_t}}\frac\partial{\partial\mathbf{h}_t}\boldsymbol{\epsilon}_t^\theta(\mathbf{x}_t,\mathbf{h}_t).
\end{equation}
which further equals to:
\begin{equation}
    \frac\partial{\partial\mathbf{h}_t}\mathbf{P}_t(\mathbf{x}_t,\mathbf{h}_t)= -\frac{\sqrt{1-\alpha_t}}{\sqrt{\alpha_t}}{J}_t.
\end{equation}
The above derivations show that ${J}_t$ represents the direction of the maximum value of image change.

\noindent\textbf{Local Mask Definition.} To specify the editing area and the non-editing area, we employed two approaches. The first method involves utilizing a pre-trained segmentation network to delineate multiple regions. Subsequently, we select the area of interest as the editing region and apply a mask to the non-editing area, using the Hadamard product to binarize the selected area. The second method consists of directly using a rectangular bounding box to define the area for editing, followed by the application of a mask to the non-editing region in a manner analogous to the previous approach. Compared to the first method, the second is more straightforward. Additionally, for the same region of interest, the area selected using this method is generally larger than that identified by the segmentation network. Overall, the second method tends to facilitate richer changes. Since we expect to find the semantics of the mask area, that is, to find the direction of change in the mask area, we add the mask to the noise prediction network ${\epsilon}_{t}(\mathbf{x}_t,\mathbf{h}_t)$:

\begin{equation}
    \tilde{\epsilon}_{t}(\mathbf{x}_t,\mathbf{h}_t)=\boldsymbol{\epsilon}_{t}^{\theta}(\mathbf{x}_{t},\mathbf{h}_{t})\odot\mathbf{mask}.
\label{mask_define}
\end{equation}
\begin{equation}
    {J}_t^{\mathrm{masked}}= \frac\partial{\partial\mathbf{h}_t}\tilde{\epsilon}_{t}(\mathbf{x}_t,\mathbf{h}_t).
\label{compute_mask}
\end{equation}

\renewcommand{\dblfloatpagefraction}{.8}
\begin{figure*}
    \centering
    \includegraphics[width=0.9\linewidth,height=10cm]{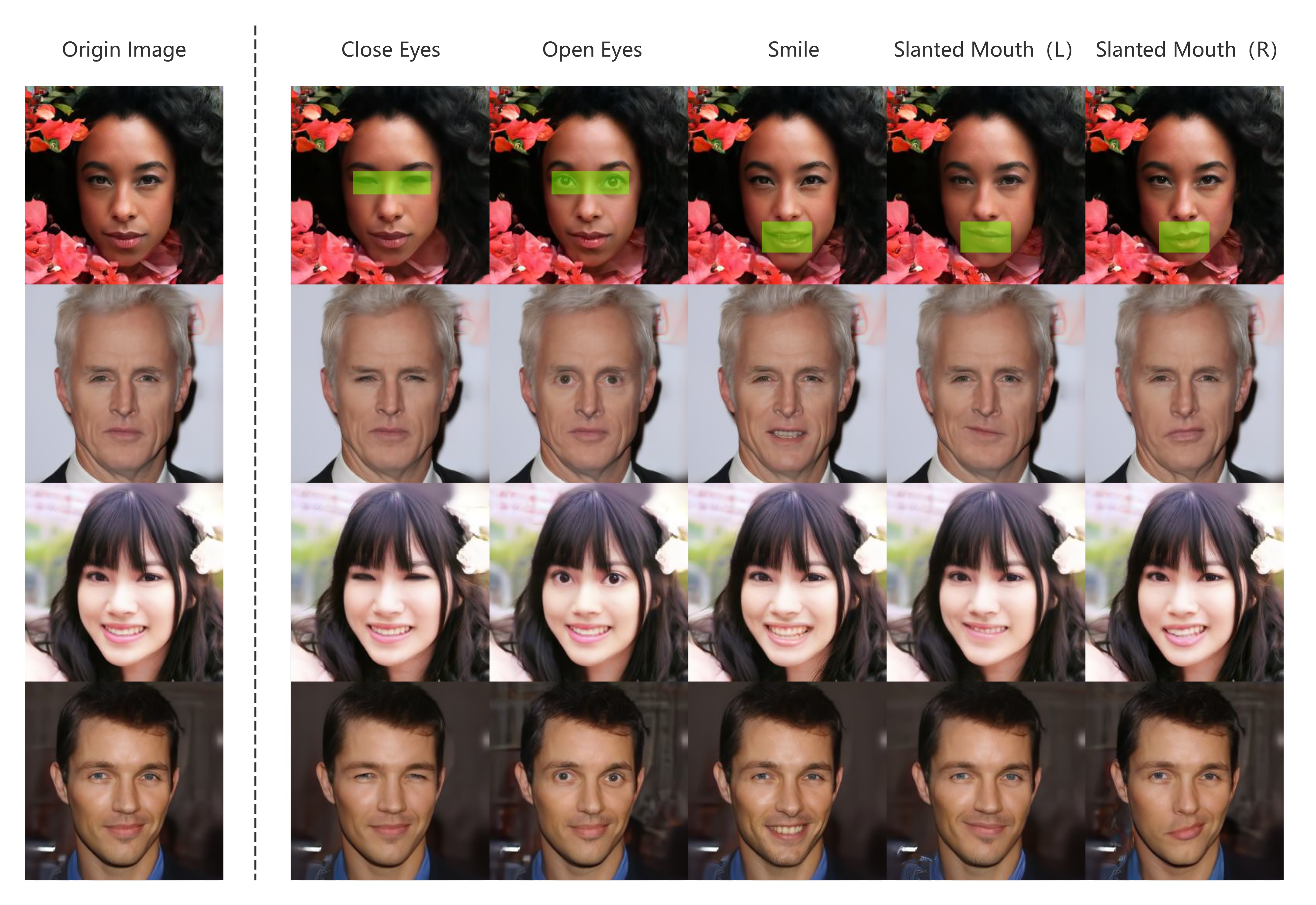}
    \caption{\textbf{Qualitative results of our method.} We experimented with the pre-trained DDPM model on the CelebA-HQ dataset with a resolution of 256*256. The leftmost image is the original image. The green box corresponds to the area where we use the mask. The area of the image mask in each column is the same. Please note that we only use the mask during training and do not need to add it during testing. In our experiment, the mask of the same area may find different attribute results. For example, for the mouth, it can be a smile or a slanted mouth in one direction.}
    \label{fig:results}
\end{figure*}
\noindent\textbf{Supressing Non-masked Semantic Variations.} Recall that our goal is to maintain the global attributes while allowing for changes in local attributes. The aforementioned algorithm identifies the vector in h-space that corresponds to the maximum change in local attributes. This approach focuses solely on local modifications, but does not address the potential for global alterations. We have previously proved that the change in x-space is equivalent to the noise prediction network $\epsilon_{t}^{\theta}(\mathbf{x}_{t},\mathbf{h}_{t})$. Therefore, our optimization goal can be expressed as:
% \begin{equation}
%     \begin{cases}\arg\max &\frac\partial{\partial\mathbf{h}_t}\tilde{\epsilon}_{t}(\mathbf{x}_t,\mathbf{h}_t)\\\\\arg\min &\frac\partial{\partial\mathbf{h}_t}{\epsilon}_{t}^{\theta}(\mathbf{x}_t,\mathbf{h}_t)\end{cases}
% \end{equation}
\begin{equation}
\arg_{\mathbf{h}_t}\max\Big(\frac\partial{\partial\mathbf{h}_t}\tilde{\epsilon}_{t}(\mathbf{x}_t,\mathbf{h}_t)\Big)\min\Big({\frac\partial{\partial\mathbf{h}_t}{\epsilon}_{t}^{\theta}(\mathbf{x}_t,\mathbf{h}_t)}\Big).
\label{eq:optimal goal}
\end{equation}

Here ${\epsilon}_{t}^{\theta}(\mathbf{x}_t,\mathbf{h}_t)$ refers to the noise prediction network in the unmasked region.

As previously discussed, changes in h-space affect the extent of modifications in $\mathbf{x}$. If a vector $h_1$ in h-space exhibits the largest change within the masked area but also significantly affects the non-masked area, we can find an alternative vector $h_2$  in h-space that induces the largest change specifically in the non-masked area. To address this, we need to consider the optimization objective in terms of Equation \ref{eq:optimal goal}.

The Jacobian matrix $J$ describes how a change in the h-space vector influences changes in $\mathbf{x}$. Therefore, $J_m$ captures the sensitivity of the masked area, and $J_u$ captures the sensitivity of the non-masked area. By leveraging these Jacobian matrices, the orthogonal projection of $h_2$  onto the non-masked area direction helps refine $h_1$ such that it optimizes the changes in the masked area without causing unintended modifications elsewhere.
Therefore,  we perform the following orthogonal Jacobian projection to achieve the optimization goal:

\begin{equation}
    J = J_{m}-\frac{J_{m}\cdot J_{u}}{J_{u}\cdot J_{u}}\cdot J_{u}.
\label{projection}
\end{equation}
Where $J_m$ and $J_u$ stand for the Jacobians of the masked and unmasked areas, respectively.

\subsection{Summary}
With the above analysis and discussion, we determine the space h-space where we are looking for semantics. We also define a new function $f$ determined by the noise prediction network $\epsilon_{t}^{\theta}$, the Jacobian matrix and its calculation method, and give a mask adding method. Solve the optimization target Equation~\ref{eq:optimal goal} by orthogonal projection (Equation \ref{projection}), so as to effectively obtain the semantic vectors of the target region.

\section{Experiments}
Since our method focuses on unsupervised semantic discovery and local image editing of diffusion models, there are not many baselines. We therefore mainly compare our method with supervised semantic discovery methods such Asyrp, Boundary Diffusion which requires attribute labels as priors, and Semdiff which mainly searches for semantic attributes globally.
\begin{table}[h]
\centering
\resizebox{85mm}{14mm}{
\begin{tabular}{lcccc}
\toprule
methods     & FID$\downarrow$   & ID$\uparrow$     & MSE($\times10^{-4}$)$\downarrow$     & LPIPS$\downarrow$   \\ \midrule
close-eyes     & 11.53 & 0.8507 & 5.86  & 0.0346  \\ 
open-eyes      & 11.50 & 0.8418 & 6.06 & 0.0312   \\ 
% mustache       & 11.61 & 0.9009 & 4.13 & 0.0251   \\ 
smile     & 14.22 & 0.8627 & 5.74 & 0.0289  \\ 
Slanted Mouth1 & 10.08 & 0.9064 & 4.51 & 0.0238   \\ 
Slanted Mouth2 & 10.30 & 0.9052 & 4.62 & 0.0231   \\ \bottomrule
\end{tabular}
}
\caption{\textbf{Quantitative performance} of our method under different attributes editing conditions}
\label{table1}
\end{table}

\subsection{Experimental Setup}
We conduct extensive experiments on multiple datasets and diffusion models with different structures to demonstrate the effectiveness of our approach. We conduct experiments on the datasets CelebA-HQ~\cite{liu2015faceattributes}, LSUN-church~\cite{yu2015lsun}, LSUN-bedroom~\cite{yu2015lsun}, and the diffusion model architectures DDPM and iDDPM~\cite{nichol2021improved}. For the comparison methods Asyrp and Boundary Diffusion, we use the checkpoints provided by the official. For the method Semdiff, since they did not provide official checkpoints, we trained it exactly as they did and found comparable properties, and then performed a fair comparison with our method. We mainly use Fréchet Inception Distance(FID)~\cite{heusel2017gans}, Mean Squared Error (MSE)~\cite{mathieu2015deep}, Identity loss (ID)~\cite{deng2019arcface} and LPIPS~\cite{zhang2018unreasonable} as objective indicators. We use FID mainly to reflect the quality of image generation. For MSE, we not only measure the global MSE, which represents the overall change of the image but also measure the MSE of the masked area and the non-masked area, which represents the local change and Changes other than local areas. We use ArcFace net to measure the face identity similarity (ID) metric. This metric describes how similar the identity of the person after the modification is to the original image, reflecting the quality of the editing. The overall structure of our experiment is as follows. We randomly select 500 images on Celeba-HQ for testing. In the first part, we show the semantic directions found by our method in the CelebA-HQ dataset. In the second part, we compare our method with three baseline methods. In the third part, we compare the editing results of our method with Semdiff in local and global regions. In the fourth part, We summarize the experimental results and analyze the advantages and disadvantages. All our experiments can be performed on a single RTX 3090 GPU.

\begin{figure}[h]
    \centering
    \includegraphics[width=0.8\linewidth]{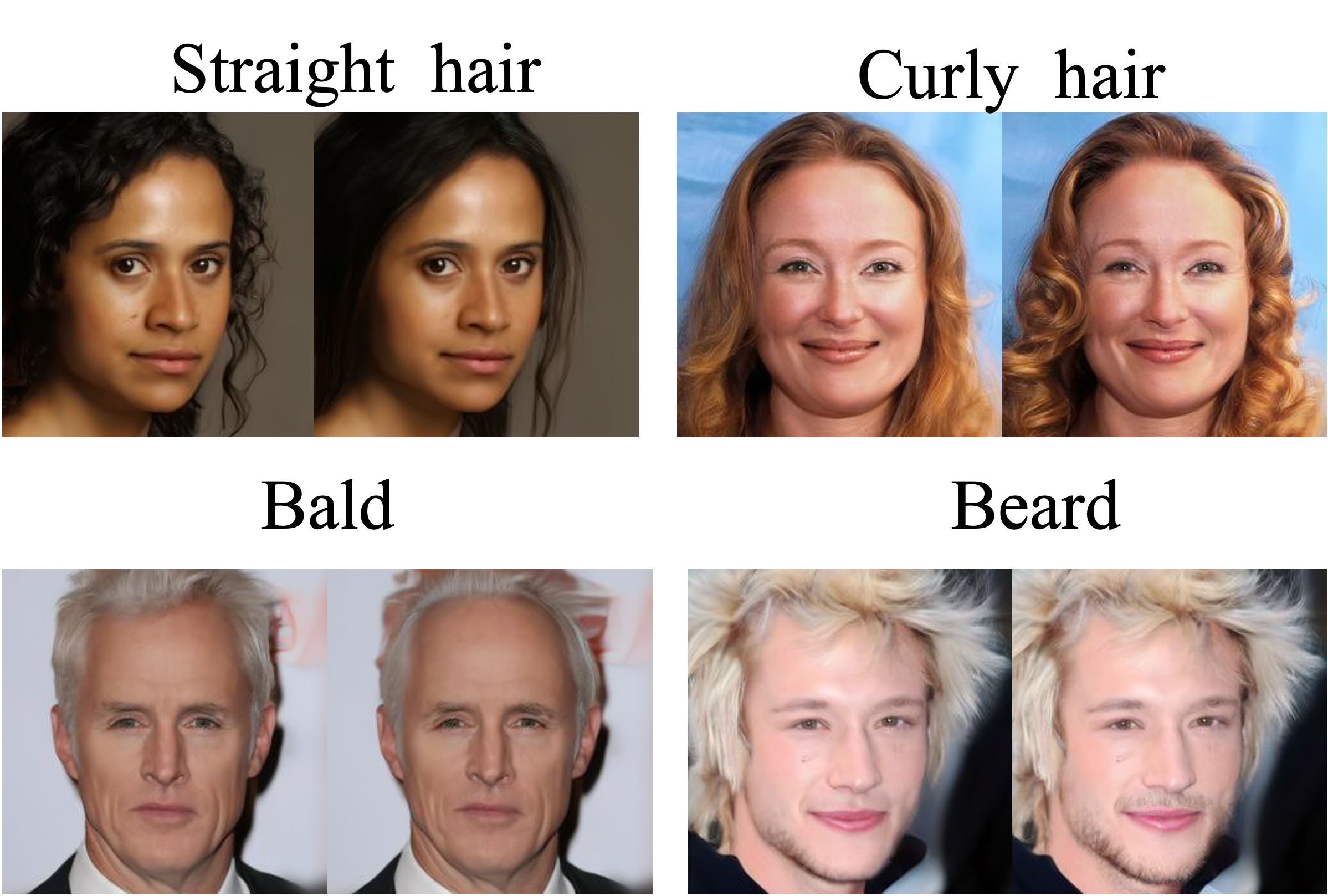}
    \caption{\textbf{More semantic editing results}}
    \label{fig:other attri}
\end{figure}

\subsection{Main Results of Attribute Editing}
In this subsection, we present the overall performance of our approach.
Our specific experimental process is as follows: First, add a mask to a specific area on the image according to the Equation \ref{mask_define}. Then we compute the Jacobian matrix of the mask and non-mask areas through Equation \ref{compute_mask}. After that, we use Equation \ref{projection} to get semantic vectors $J$. Finally, We apply the Jacobian matrix in timestpes of reverse process to obtain the results.

We mainly select the eyes region and the mouth region to find semantics. In Figure \ref{fig:results}, we show the semantics found by our method in the eye and mouth regions. In the eye semantics, there are eyes open and eyes closed; in the mouth attributes, there are smiles, and slanting to the left and right. Interestingly, we also found some semantics for specific groups of people, such as editing the beard area for men, and changing the hair of people with long curly hair to straight hair. We show the above in Figure~\ref{fig:other attri} and put more semantic editing results of other regions in the Appendix.

We further test the quantitative results shown in Table \ref{table1}. Our results show that while achieving good editing results, the image quality indicators FID, MSE, LPIPS, and the identity change ID of the person have achieved extremely high levels. 
\begin{table}[h]
\centering
\resizebox{85mm}{12mm}{
\begin{tabular}{lcccc}
\toprule
methods     & FID$\downarrow$   & ID$\uparrow$     & MSE($\times10^{-4}$)$\downarrow$     & LPIPS$\downarrow$   \\ \midrule
Asyrp   & 54.04 & 0.4892 & 16.60 & 0.2172 \\ 
BoundaryDffusion & 53.04 & 0.5671 & 15.14 & 0.1765 \\ 
% baseline-smile & 16.60 & 0.7459 & 8.29 & 0.0424 \\ 
Semdiff &\underline{21.77} &\underline{0.7224} &\textbf{5.23} &\underline{0.0424} \\ 
ours   &\textbf{14.22} &\textbf{0.8627} &\underline{5.74} &\textbf{0.0289}  \\ \bottomrule
\end{tabular}
}
\caption{\textbf{Quantitative comparison} between different local editing approaches and ours. We compare the editing effects of the attribute 'Smile'.  \textbf{Bold} and \underline{underline} represent the best and second-best performance, respectively.}
\label{table2}
\end{table}

\subsection{Comparison against Other Baselines}
In this section, we compare our method with the state-of-the-art methods both qualitatively and quantitatively. We compare with Asyrp, Boundary diffusion, and Semdiff, which are current methods for semantic discovery and editing using diffusion models. 
\begin{figure}[t]
    \centering
    %\hspace{-0.5cm}
    \includegraphics[width=0.9\linewidth,height=8cm]{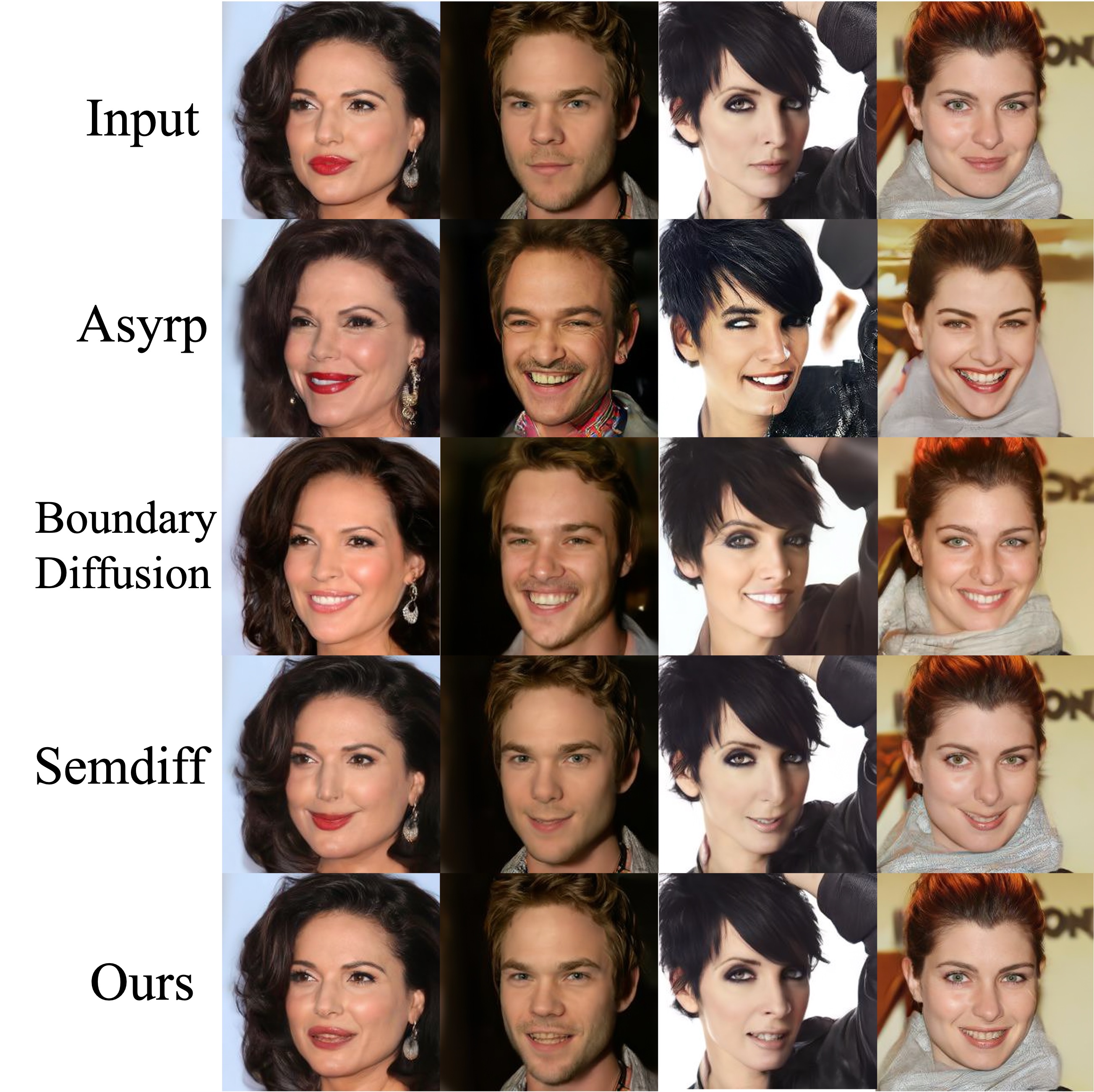}
    \caption{\textbf{Qualitative results.} Our method has the best results on the overall structure and details.}
    \label{fig:results against pipeline}
\end{figure}
Although our method finds local semantics, the smile attribute exists in the semantics of the mouth. And for fair comparison, their method has an open source smile attribute checkpoint, so we used the smile attribute that is unfavorable to ours for comparison. But we are surprised to find that even so, our unsupervised method is better than the supervised method, and has a great improvement effect than the unsupervised semdiff.

In Figure \ref{fig:results against pipeline}, we show the qualitative results. As can be seen from Figure \ref{fig:results against pipeline}, Asyrp will bring about a great change in the image structure, and for images with complex character postures, it will bring more serious distortion results. Boundary Diffusion is better than Asyrp, but it still has a more serious impact on character identity, background and character details. The effect of Semdiff is intuitively good, but because this method only considers local semantic attributes, it does not consider the influence of other global parts. Therefore, the details of the characters, such as the noses in the first and third images, and the beards in the second male image, have more serious changes, which is not desirable. The last row is the result of our method. Our method not only ensures the invariance of the overall image (character identity, background, etc.), but also ensures the invariance of character details (clothes, nose, beard, etc.).
\begin{figure}[t]
    \centering
    \includegraphics[width=1\linewidth]{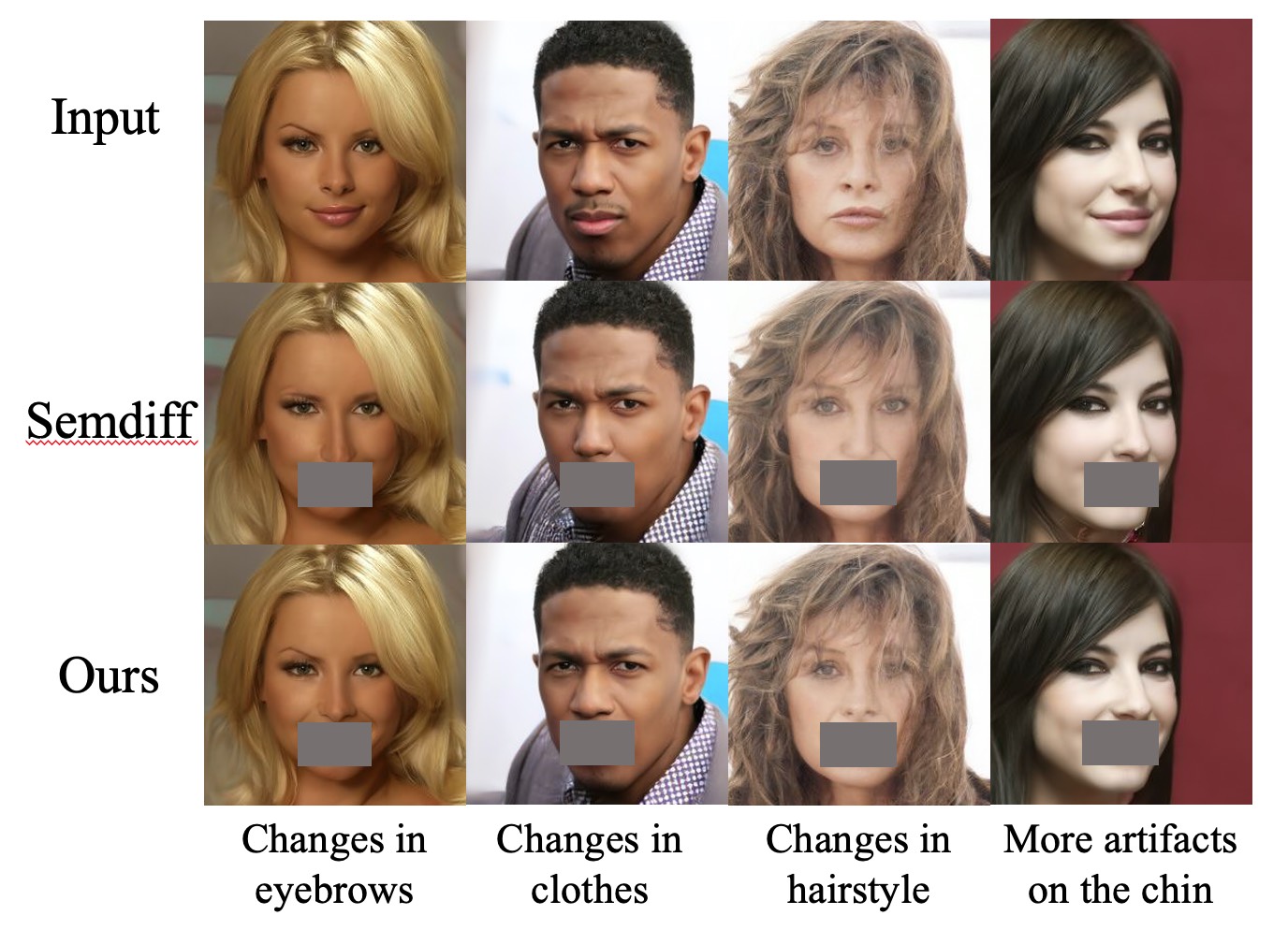}
    \caption{\textbf{Exemplary comparison of Semdiff and our RBE}. We compare the remaining parts after editing the mask of the region. Our method has a clear advantage in image preservation outside the region. }
    \label{fig:comparision}
\end{figure}

Table \ref{table2} presents a quantitative evaluation of baselines and our method using metrics such as FID, ID, MSE, and LPIPS. Our method delivers best results across most metrics, particularly excelling in FID and ID, which assess image quality and person identity, respectively. MSE reflects image quality on the one hand and editing intensity on the other. Asyrp and Boundary Diffusion exhibit significantly higher MSE values due to larger structural changes. Although our method and Semdiff have slightly higher in MSE values, it does not necessarily indicate inferiority, as other metrics should also be considered for a comprehensive assessment.

\begin{figure}[t]
    \centering
    \includegraphics[width=0.95\linewidth]{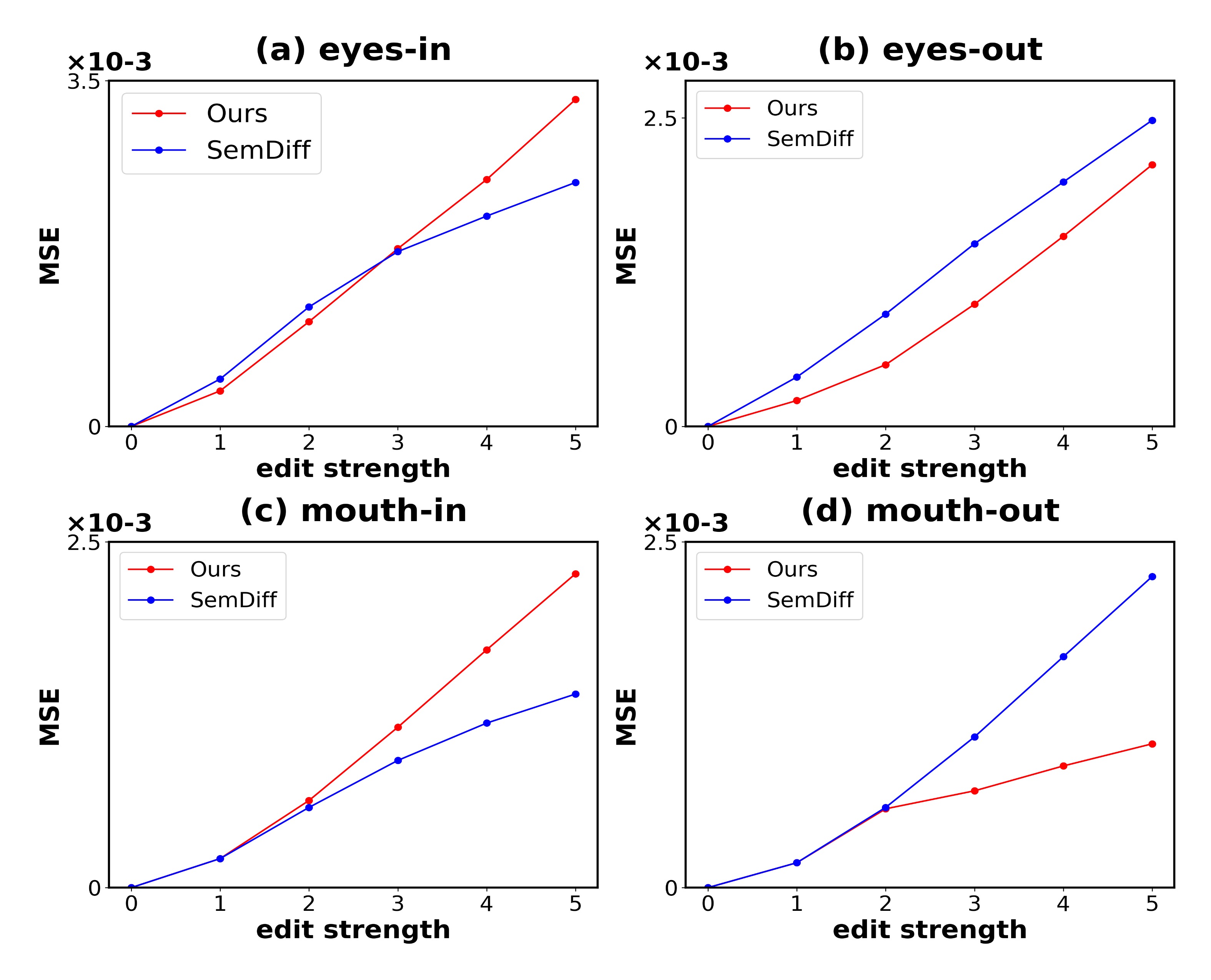}
    \caption{\textbf{Quantitative results of pixel change} in mouth editing and eyes editing. Our method is shown in red and Semdiff is shown in blue. "In" and "out" refer to the region of interest and its surroundings, respectively. For effective local editing, a higher change within the region of interest ("in") and a lower change in the surrounding area ("out") are expected.}
    \label{fig:Quantitative results of pixel change}
\end{figure}
\subsection{Changes of Local and Non-local Regions}
To demonstrate the superiority of our method in local region editing, we compared the degree of change in local and non-local regions with Semdiff. we compared the MSE inside and outside the mask region when editing the eyes and mouth. Compared with Semdiff, when the overall MSE scale is close, our method is more inclined to edit the inner area of the mask, and the changes to the outer area of the mask are less than Semdiff. The qualitative results in Figure~\ref{fig:Quantitative results of pixel change} also prove this point: when editing the masked region, Semdiff will cause changes in the rest of the image, while our method maintains the non-masked region well.

\subsection{Discussion}
We have demonstrated the state-of-the-art performance of our method in local attribute editing. More experimental results, including other attribute edits and different datasets, are available in the Appendix. First, since our method is aimed at local editing, when editing global attributes, the mask area needs to be expanded to a large enough area. Under this premise, our method will not be significantly better than previous research on global attribute editing. Moreover, when the mask area is very small (such as an eye, a tooth, etc.), our method cannot find significant semantic directions. Our future research will be directed towards extending our method to more fine-grained areas and global attributes.

\section{Conclusion}

To explore the latent potential within diffusion models, we propose a novel method utilizing the Jacobian projection technique for precise semantic editing of pre-trained models without additional training. Our approach allows for the identification and manipulation of semantic attributes in local regions by projecting the Jacobian of the masked area into a low-dimensional subspace orthogonal to the non-masked regions. This technique significantly enhances control over local image attributes while preserving overall image harmony. Furthermore, our method outperforms existing supervised approaches, especially in specific tasks such as facial attribute editing, demonstrating its substantial potential to advance unsupervised image editing techniques.

\section{Acknowledgment}
This work was supported in part by the National Key R\&D Program of China (No.2021ZD0112100), Fundamental Research Funds for the Central Universities (No.2022XKRC015), National NSF of China (No.U24B20179, No.62120106009, No.62372033) and Beijing Natural Science Foundation (L242021).
\bibliography{ref}

\end{document}